\documentclass[conference]{IEEEtran}
\IEEEoverridecommandlockouts

\usepackage{cite}
\usepackage{amsmath,amssymb,amsfonts}
\usepackage{algorithmic}
\usepackage{graphicx}
\usepackage{textcomp}
\usepackage{xcolor}
\usepackage{booktabs}
\usepackage{multirow}
\usepackage[hyphens]{url}
\def\BibTeX{{\rm B\kern-.05em{\sc i\kern-.025em b}\kern-.08em
    T\kern-.1667em\lower.7ex\hbox{E}\kern-.125emX}}

\usepackage{wrapfig}
\usepackage{hyperref}

\begin{document}

\title{Exploring the Impact of Dataset Statistical Effect Size on Model Performance and Data Sample Size Sufficiency}


\author{%
\IEEEauthorblockN{%
Arya Hatamian\IEEEauthorrefmark{2},
Lionel Levine\IEEEauthorrefmark{3}$^*$,
Haniyeh Ehsani Oskouie\IEEEauthorrefmark{3}$^*$ \thanks{$^*$ Equal contribution.},
Majid Sarrafzadeh\IEEEauthorrefmark{3}}
\IEEEauthorblockA{\IEEEauthorrefmark{2}\textit{Department of Business Administration and Management}, University of California, Riverside\\}
\IEEEauthorblockA{\IEEEauthorrefmark{3}\textit{Department of Computer Science}, University of California, Los Angeles\\
ahata006@ucr.edu}
}

\maketitle

\begin{abstract}
Having a sufficient quantity of quality data is a critical enabler of training effective machine learning models. Being able to effectively determine the adequacy of a dataset prior to training and evaluating a model's performance would be an essential tool for anyone engaged in experimental design or data collection. However, despite the need for it, the ability to prospectively assess data sufficiency remains an elusive capability. We report here on two experiments undertaken in an attempt to better ascertain whether or not basic descriptive statistical measures can be indicative of how effective a dataset will be at training a resulting model. Leveraging the effect size of our features,  this work first explores whether or not a correlation exists between effect size, and resulting model performance (theorizing that the magnitude of the distinction between classes could correlate to a classifier's resulting success). We then explore whether or not the magnitude of the effect size will impact the rate of convergence of the learning curve, (theorizing again that a greater effect size may indicate that the model will converge more rapidly, and with a smaller sample size needed). Our results appear to indicate that this is not an effective heuristic for determining adequate sample size or projecting model performance, and therefore that additional work is still needed to better prospectively assess adequacy of data.
\end{abstract}

\textbf{Code:} \href{https://colab.research.google.com/drive/1-XdXXZz3bYmxBner_FHdnKOKcMTrnI8s?usp=sharing}{Colab Notebook 1} \href{https://colab.research.google.com/drive/1EJ8wle77zXTHhnYgtkXCU2-fRSsmGUC9?usp=sharing}{Colab Notebook 2}

\subsection{Introduction}
The sufficiency of a dataset, both in terms of size and representativeness, is critical to training an effective Machine Learning model. Failure to train on data of adequate quality is likely to result in models that dramatically under-perform in production environments \cite{Friedland2024}. 

Training machine learning models on inadequate data presents several challenges that can significantly impact model performance and generalizability. While the problem of 'inadequacy' of data is a multifaceted one, one significant component of it is a lack of sufficient data volume, which in turn likely impacts the representativeness of the data sample towards the overall population undermining the ability of a model to learn meaningful patterns, resulting in overfitting, where the model performs well on the training data but struggles to generalize to unseen data leading to poor outcomes when deployed in real-world scenarios. This owes to the fact that if the dataset used for training does not accurately reflect the conditions and distributions the model will encounter in real-world applications, the model will struggle to generalize beyond the narrow scope of its training \cite{9517732}. Consequently, the model's assumptions about the domain may be incorrect, leading to suboptimal decision-making in practical applications.

In cases where the dataset is particularly small or inadequate, machine learning models—especially complex ones like deep learning architectures—fail to learn the intricate relationships within the data. Complex models require large amounts of data to effectively capture subtle patterns \cite{keshari2020unravellingsmallsamplesize}. Without sufficient data, the model may not converge, or it may underfit, meaning it performs poorly on both the training and test sets due to an inability to learn the underlying patterns. 
Moreover, the use of inadequate data increases variability in model performance. Small datasets are more sensitive to changes in initialization values, training procedures, or even small differences in the data samples used during training \cite{dodge2020finetuningpretrainedlanguagemodels}. As a result, models trained on inadequate data exhibit inconsistent and unpredictable behavior, making them unreliable in different testing or production environments. Issues such as overfitting, bias toward dominant classes, poor handling of noisy data, and a lack of generalizability can result from insufficient, non-representative, or poor-quality data. Ensuring access to high-quality, diverse, and sufficiently large datasets is essential for developing models that are both robust and reliable in their predictions.

Having a reliable means of determining the quantity of training data required to effectively train a model would be an incredibly useful tool for researchers to have. This owes both to the upfront challenges and associated costs of data collection, and with the advent of increasingly costly training runs, the investment costs associated with training.
This is a widely recognized need across research domains, with many experimental protocols requiring power analyses, or equivalent statistical studies to prospectively determine the requisite amount of data collection required to run studies effectively.\cite{Bausell_Li_2002}
However, in spite of the obvious utility of such an assessment, it remains elusive for machine learning model developers. As already noted, many methods for external dataset-based validation but are post-hoc model dependent.

\subsection{Background}
\subsubsection{Prospective analysis of data sufficiency}
In most academic disciplines, conducting a sufficiently robust study design is arguably more important than the statistical analysis that succeeds it. After-all, a poorly designed study may be unsalvageable after the fact, whereas a poorly analyzed study can simply be re-analyzed once errors in methodology are determined \cite{bmj_study_design}. A critical component in study design is the determination of the appropriate sample size. The sample size must be large enough to establish statistical significance of any potential findings, yet not so large as to unnecessarily burden researchers with the (often costly) acquisition of data. Attempting to determine data sufficiency for machine learning (ML) models is a particularly vexing challenge given the nature of machine learning. Unlike statistics, ML does not attempt to assert any factual relationship between label and feature set \cite{bennett2022similaritiesdifferencesmachinelearning}; Rather ML models are almost mercenary in nature. Charged with best accomplishing a given prediction or classification task, irrespective of any actual underlying statistical relationship. 
This relaxed focus allows for a more expansive inclusion of features for which a statistically significant relationship to the label may not exist.

\subsubsection{Current approaches to determining sufficient sample size}
There is a lack of approaches that exist for determining a sufficient sample size to train ML models. The absence of a robust pre-hoc approach that is broadly applicable remains a significant challenge in machine learning, largely stemming from the limited research conducted in this area \cite{BALKI2019344}.
The most common current approach is empirical analysis. This post-hoc method occurs when the sample size for training is incrementally increased and the learning curve is analyzed to see where the point of diminishing returns occurs in relation to sample size. This enables the identification of the correlation between sample size and model performance. The only existing approaches that can be considered a pre-hoc approach are model-based. These approaches utilize the parameters of an algorithm to determine sample size. Some models, like neural networks, typically need more data to find a pattern and develop a function that maximizes accuracy when compared to simpler models like decision trees. This is due to the nature of how the model is developed to predict. The more parameters a model has, the more complex it tends to be. Neural networks typically have multiple parameters in each of the hidden layers whereas other algorithms only have one layer. There are mathematical equations like those of Baum and Hausler that allow one to determine a sufficient number of samples through this model-based framework. However, this framework is not optimal for all tasks and has its trade-offs.\cite{BALKI2019344}

\subsubsection{Statistical effect size}
Effect size is a quantitative measure of the magnitude of a phenomenon. It provides an indication of the practical significance of a result, independent of sample size, making it a critical complement to p-values in hypothesis testing. While p-values indicate whether a result is statistically significant, effect size reveals how strong the effect is, giving a clearer sense of its importance in real-world terms \cite{10.4300/JGME-D-12-00156.1}.
In categorical classification modeling challenge, effect size can serve as a critical pre-training tool to compare differences in feature space between classes. 

Cohen's d is commonly used for continuous feature variables, while odds ratios (ORs) are suitable for categorical variables. Both measures help quantify the magnitude of differences across populations. The formula is
\(
    d = \frac{\mu_1 - \mu_2}{SD_{\text{pooled}}},
\)
where \(\mu_1\) and \(\mu_2\) are the means of the two populations, and the pooled standard deviation (\(SD_{\text{pooled}}\)) is calculated as
\(
        SD_{\text{pooled}} = \sqrt{\frac{(n_1 - 1) \cdot SD_1^2 + (n_2 - 1) \cdot SD_2^2}{n_1 + n_2 - 2}}.
\)

The odds ratio (OR) measures the strength of association between a binary categorical variable (e.g., presence/absence of a condition) and group membership. The formula is
\(
    OR = \frac{\text{Odds in Group 1}}{\text{Odds in Group 2}}.
\)
Odds are calculated as the ratio of the probability of an event occurring to the probability of it not occurring
\(
    \text{Odds} = \frac{p}{1 - p},
\)
where \(p\) is the probability of the event. This allows us to calculate the individual effect sizes for each feature against a class \cite{doi:10.5395/rde.2015.40.4.328}. The cumulative effect size of an entire dataset against a class, can then be averaged for a resulting cumulative effect size score.

While these univariate metrics provide insight as to how significant each feature is to separating between classes, they do not capture the multivariate structure of the data. Machine learning models optimize objective functions by operating on the entire feature vector, which includes analyzing the relationships between features of different coordinates within these vectors. To ensure that this is captured in our effect size calculation we also include a Mahalanobis separation. This is utilized as a multivariate measure of class separability and can be calculated as
\(
D_M = \sqrt{(\mu_1 - \mu_2)^T \Sigma^{-1} (\mu_1 - \mu_2)}.
\)

\subsection{Hypothesis}
This research attempts to ascertain whether or not certain descriptive statistical features of a dataset can be indicative of prospective model performance, and can provide a heuristic for data volume required to achieve certain generalizable performance benchmarks.

\textbf{Corollary:} The magnitude of Feature Effect size, and data volume, both impact model performance and generalizability, with a higher effect-size offsetting the need for large datasets, while a smaller effect size generally requiring a greater volume of data to achieve similar results.

\textbf{Corollary:} There is an upper-bound to performance irrespective of data size, and possibly effect size (i.e., as data size goes to infinite model performance plateaus). 

\begin{figure}[t]
    \centering
    \includegraphics[width=0.45\textwidth]{fig_111_adult_avg.pdf}
    \caption{Experiment 1.1.1 (income as label): Mean absolute effect size plotted against model performance (accuracy, precision, recall, and F1-score). Each point represents a model (distinguished by color) trained on a distinct subset.}
    \label{fig:exp111_income}
\end{figure}

\begin{figure}[t]
    \centering
    \includegraphics[width=0.45\textwidth]{fig_111_bank.pdf}
    \caption{Experiment 1.1.1 (deposit subscription as label): Multivariate effect size plotted against model performance (accuracy, precision, recall, and F1-score). Each point represents a model (distinguished by color) trained on a distinct subset.}
    \label{fig:exp111_deposit}
\end{figure}

The goal of this research is as follows: If a researcher has advanced knowledge the size of the dataset employable for training, and can calculate the effect size of features with respect to the labels, one can make a reasonable assumption on projected model performance. 

To best explore this overall objective, we conducted two specific experiments to determine whether or not such a relationship exists.

\textbf{Hypothesis 1:} There is a correlation between statistical effect-size and model performance up to a point. 

\textbf{Hypothesis 2:} There is a correlation between statistical effect-size, data size, and model performance up to a point.

These trends are universal and applicable across datasets.

\section{Methods}
We used two tabular datasets from the UCI Machine Learning Repository: Adult (Census Income) with 48,842 instances and 14 features \cite{adult_2}, and Bank Marketing with 45,211 instances and 16 features \cite{bank_marketing_222}.
Opting for a binary classification problem, the label under study was income for the Adult Cencus data, which was segmented into a binary categorical segmented as either greater than 50k or less than or equal to 50k per year. For the Bank data, the label was whether each client subscribed a term deposit (yes or no).

\begin{figure}[t]
    \centering
    \includegraphics[width=0.45\textwidth]{fig_112_adult_avg.pdf}
    \caption{Experiment 1.1.2 (gender as label): Mean absolute effect size plotted against model performance (accuracy, precision, recall, and F1-score). Each point represents a model (distinguished by color) trained on a distinct subset.}
    \label{fig:exp112_gender}
\end{figure}

\begin{figure}[t]
    \centering
    \includegraphics[width=0.45\textwidth]{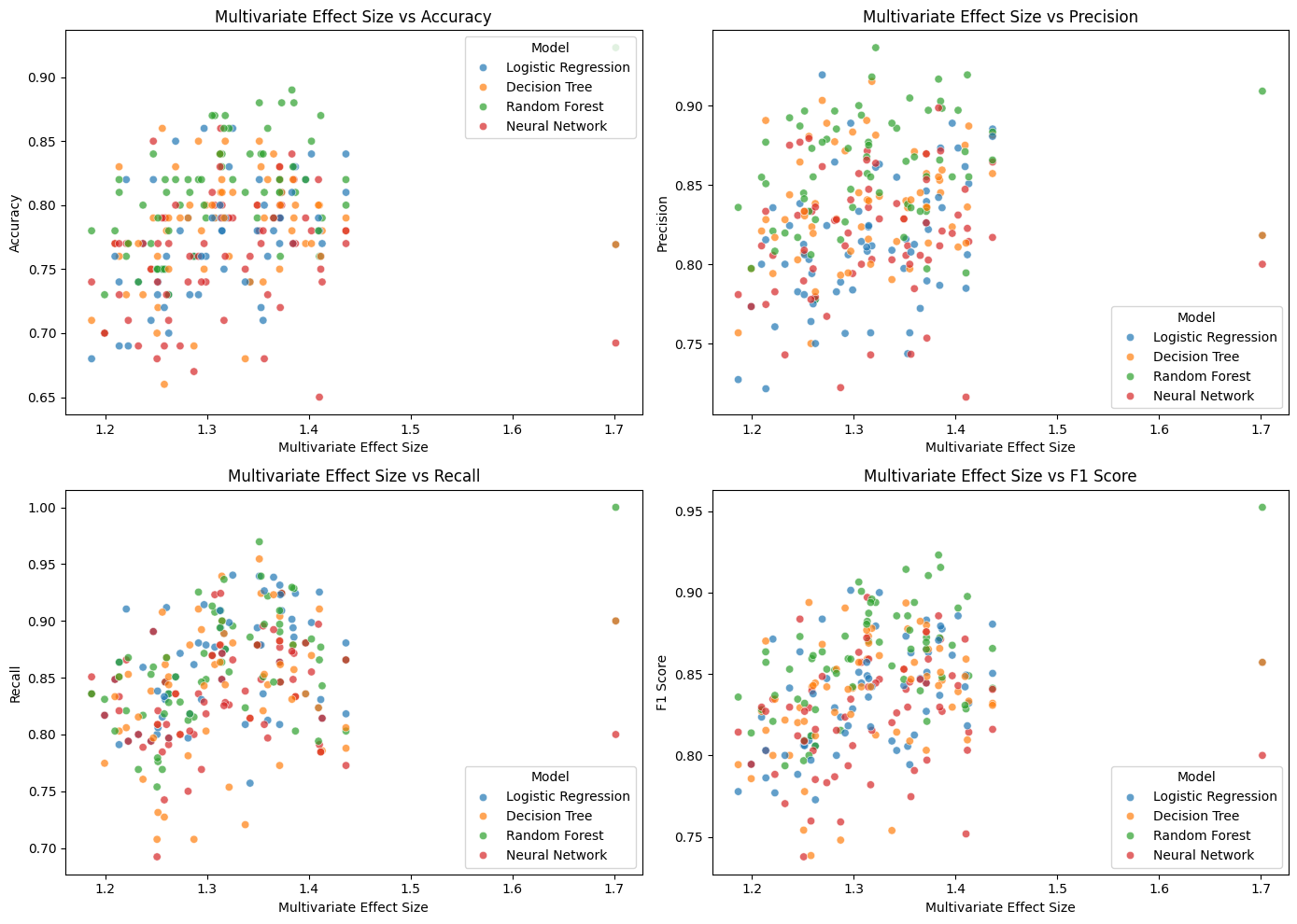}
    \caption{Experiment 1.1.2 (gender as label): Multivariate effect size plotted against model performance (accuracy, precision, recall, and F1-score). Each point represents a model (distinguished by color) trained on a distinct subset.}
    \label{fig:exp112_sex}
\end{figure}

\begin{figure}[t]
    \centering
    \includegraphics[width=0.45\textwidth]{fig_121_adult_avg.pdf}
    \caption{Experiment 1.2.1 (income as label): Effect size of removed feature plotted against model performance drop (accuracy drop, precision drop, recall drop, and F1-score drop). Each point represents a model (distinguished by color) with a different feature removed.}
    \label{fig:exp121_income}
\end{figure}

\begin{figure}[t]
    \centering
    \includegraphics[width=0.45\textwidth]{fig_122_adult_avg.pdf}
    \caption{Experiment 1.2.2 (gender as label): Effect size of removed feature plotted against model performance drop (accuracy drop, precision drop, recall drop, and F1-score drop). Each point represents a model (distinguished by color) with a different feature removed.}
    \label{fig:exp122_gender}
\end{figure}

\begin{figure}[t]
    \centering
    \includegraphics[width=0.45\textwidth]{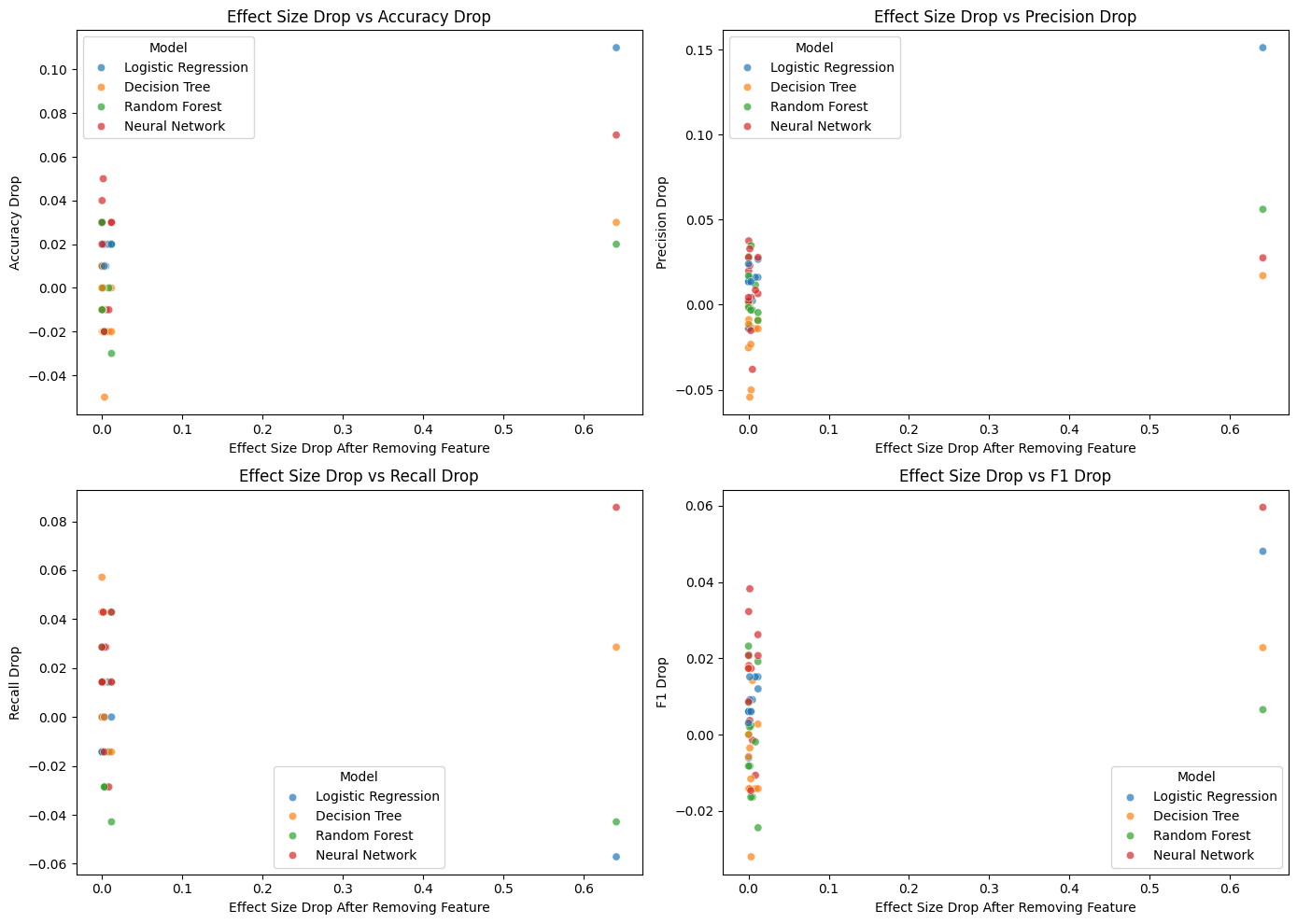}
    \caption{Experiment 1.2.2 (gender as label): Multivariate effect size drop plotted against model performance drop (accuracy drop, precision drop, recall drop, and F1-score drop). Each point represents a model (distinguished by color) with a different feature removed.}
    \label{fig:exp122_sex}
\end{figure}

\begin{figure}[t]
    \centering
    \includegraphics[width=0.42\textwidth]{fig_2122_adult_avg.pdf}
    \caption{Experiment 2.1 (left): Mean absolute effect size plotted against slope of the learning curve. Experiment 2.2 (right): Mean absolute effect size plotted against slope of the difference between training learning curve and validation learning curve. Both experiments set income as label. Each point represents a model (distinguished by color) trained on a distinct subset.}
    \label{fig:exp21}
\end{figure}

Standard preprocessing of the data was undertaken for both datasets, including removing rows with null values, encoding categorical variables, and standardizing the feature space. Categorical features were encoded using LabelEncoder, converting each category into a numerical value. All numerical features were standardized using StandardScaler.  The standardized numerical features and encoded categorical features were combined into a single feature matrix, which was then used for training and evaluating the machine learning models. 

The following set of experiments was then performed on both datasets:

\subsubsection{Experiment 1} \textbf{Determine if effect size correlates to performance across datasets and feature mixes.}

\textbf{Experiment 1.1: Different subsets of data, with the same set of Features, (i.e., different mix of rows within the same dataset)}

For this experiment, each dataset was segmented into subsets, each with 500 rows. All features remained identical across subsets, with only individual rows varying across datasets.

\textbf{Experiment 1.1.1:} Model performance was compared across the same set of features/different data. For this experiment, the "Income Greater Than 50k" binary categorical feature served as label for the Adult Census data, with the rest of the features serving as the feature set. For the Bank data, whether the client subscribed to a term deposit served as the binary label. For the Adult Census data, the average effect size was calculated for each subset. The multivariate effect size (Mahalanobis separation) was calculated for subsets of both datasets for the binary labels we will aim to classify. 
The effect size for numerical features was calculated using Cohen's d, which measures the standardized difference between the means of two groups (e.g., individuals with income greater than \$50K and those with less than or equal to \$50K). For categorical features, the effect size is calculated using the odds ratio derived from a contingency table that compares the frequency distribution of the categories across the target variable (income).
After calculating the effect size for each individual feature, the average effect size across all features within a subset is computed. This average provides a single summary metric representing the overall effect size of the dataset in that particular subset. This was only done for the Adult Census data.
The multivariate effect size measure for each subset was calculated with the distance function as described previously for both datsets.  

An identical sequence of classifiers was trained on the models, employing different classification techniques. This was done to try and generalize the approach across learning model paradigms. Specifically, for every experiment, we employed the same four machine-learning models: Logistic Regression, Decision Tree, Random Forest, and Neural Network. This yielded data points of model performance vs. averaged effect size and model performance vs. multivariate effect size, allowing us to compare those two elements by calculating a direct correlation. 

\textbf{Experiment 1.1.2:} Attempted to employ a different feature mix on the same rows, (i.e., utilizing the same hundred subsets of data but with a different set of features and labels). Specifically the income label was reintroduced into the feature-mix and this time the "sex" binary feature was employed as the label for the Adult Census data. And a similar approach for the Bank data, where the label became whether the client has a housing loan. The same set of models were run and performance metrics calculated.

\textbf{Experiment 1.2: } \textbf{Compare model performance across a mix of features with same underlying data}

\textbf{Experiment 1.2.1:} For this experiment, model performance across a differing set of features with the same underlying data was compared. In this experiment the ”income greater than 50k” binary was retained as a label for the Adult Census dataset and whether a client subscribed a term deposit for the Bank dataset. The rest of the features were retained as a feature set.  A baseline on all performance metrics was initially calculated. For this experiment, a series of subsets were devised. For each subset one of the features was dropped, and the effect size of the dropped feature was calculated, again for the Adult Census data. An identical sequence of classifiers was then trained, and model performance drop was calculated. This produced 14 data points of model performance drop vs. effect size of removed feature. The multivariate approach calculated the baseline multivariate effect size. Then, for each subset where a different feature was dropped the multivariate effect size was calculated again. We then plotted the multivariate effect size drop vs. the performance drop for both datasets.

\textbf{Experiment 1.2.2}: In a parallel fashion to experiment 1.1.2, we reran 1.2.1, but this time used the ”sex” binary label as a feature for the Adult Census data and "housing" for the Bank data. An additional set of data points were generated.

\subsubsection{Experiment 2} \textbf{Compare the relationship between effect size and learning curve slope.}

This experiment sought to determine if there is a relationship between the slope of a learning curve (the rate at which a model’s error rate decreases with respect to training data size) and the effect size, irrespective of overall model performance. The hypothesis being tested was whether or not a more dramatic effect size would indicate a cleaner dataset, requiring fewer samples for a model to ‘bottom out’ in terms of error. The critical distinction here was that it would not necessarily predict model performance, but rather how large a dataset was required for convergence on an optimal model. For each model in experiment 1, where a previously calculated effect size was obtained, a learning curve/error rate for both training and validation sets was plotted.

\textbf{Experiment 2.1: } \textbf{Correlation of effect size to slope of learning curve}

For each model, an associated learning curve on a validation set was plotted. This model generally adhered to a logarithmic function with an associated coefficient, beginning elevated, but converging on an optimal error rate with the addition of more training data. This slope was then plotted against the associated effect size and the correlation computed.

\textbf{Experiment 2.2: } \textbf{Correlation of Effect Size to Ratio of Error between Training and Validation Sets}

For each model, in addition to learning curve on the validation set, a learning curve on the training set was also plotted. These models tended to start with virtually no error, as a small amount of training data allowed for significant overfitting. As more training data is introduced, the error rate increases until it converges with the validation set's error rate. Pairing both sets of plots, for each point, the magnitude of the difference in error between the training and validation sets was calculated. This resulted in a linearly decreasing amount as error rates converge. Then, take slope of this linear line representing the magnitude of error difference was then extracted and correlated against effect size. 

Both of these experiments were carried out with an average effect size for the Adult Census data and multivariate effect size for both datasets.

\section{Results}

Multiple inferential metrics were utilized to evaluate results.  Pearson correlation (r with associated p-value) was used to evaluate linear relationship. The r-squared value was derived to determine how much of the variance in the dependent variable can be explained by the independent variable. Spearman correlation ($\rho$ with associated p-value) was examined to detect monotonic relationship. Kendall’s Tau ($\tau$ with associated p-value) was computed to establish whether any consistent ordering between variables exists. A 95\% bootstrap confidence interval for Pearson and Spearman was also implemented for further validation of results, since experiments were run on a single dataset. Here, the data was resampled 1000 times to see if any correlation that does exist is sound based on if it shows up consistently. Table \ref{table:metric_summary}, \ref{table:multivariate_adult_summary}, \ref{table:multivariate_bank_summary} details the experimental results and plots for each experiment can be found in Figures \ref{fig:exp111_income}, \ref{fig:exp111_deposit}, \ref{fig:exp112_gender}, \ref{fig:exp112_sex}, \ref{fig:exp121_income}, \ref{fig:exp122_gender}, \ref{fig:exp122_sex}, and \ref{fig:exp21}.

\begin{table*}[t]
\centering
\caption{Complete Summary of Inferential Metrics For Average Effect Size Experiments — Adult Census Dataset}
\label{table:metric_summary}
\resizebox{0.9\textwidth}{!}{
\begin{tabular}{|c|c|c|c|c|c|c|c|c|c|c|c|}
\hline
Exp. & Metric &
Pearson $r$ & $p$ & $R^2$ &
Spearman $\rho$ & $p$ &
Kendall $\tau$ & $p$ &
Slope &
Pearson CI &
Spearman CI \\
\hline

\multicolumn{12}{|c|}{\textbf{Experiment 1.1.1 — Income Label, Identical Features}} \\
\hline
1.1.1 & Accuracy  & -0.083 & 0.181 & 0.007 & -0.052 & 0.399 & -0.035 & 0.417 & -0.028 & [-0.205, 0.047] & [-0.172, 0.067] \\
1.1.1 & Precision & -0.056 & 0.362 & 0.003 & -0.033 & 0.590 & -0.024 & 0.568 & -0.059 & [-0.164, 0.054] & [-0.152, 0.088] \\
1.1.1 & Recall    &  0.101 & 0.102 & 0.010 &  0.121 & 0.049 &  0.080 & 0.057 &  0.101 & [-0.013, 0.217] & [0.007, 0.241] \\
1.1.1 & F1 Score  &  0.061 & 0.327 & 0.004 &  0.078 & 0.204 &  0.049 & 0.236 &  0.051 & [-0.058, 0.171] & [-0.038, 0.198] \\
\hline

\multicolumn{12}{|c|}{\textbf{Experiment 1.1.2 — Sex Label, Identical Features}} \\
\hline
1.1.2 & Accuracy  & -0.173 & 0.005 & 0.030 & -0.150 & 0.014 & -0.101 & 0.018 & -0.072 & [-0.287, -0.056] & [-0.272, -0.038] \\
1.1.2 & Precision & -0.139 & 0.024 & 0.019 & -0.123 & 0.046 & -0.084 & 0.045 & -0.051 & [-0.254, -0.014] & [-0.238, 0.001] \\
1.1.2 & Recall    & -0.145 & 0.018 & 0.021 & -0.086 & 0.165 & -0.057 & 0.169 & -0.065 & [-0.265, -0.027] & [-0.204, 0.025] \\
1.1.2 & F1 Score  & -0.185 & 0.003 & 0.034 & -0.144 & 0.019 & -0.097 & 0.021 & -0.058 & [-0.305, -0.066] & [-0.263, -0.029] \\
\hline

\multicolumn{12}{|c|}{\textbf{Experiment 1.2.1 — Income Label, Feature Removal}} \\
\hline
1.2.1 & Accuracy Drop  & -0.123 & 0.365 & 0.015 & -0.003 & 0.982 & -0.017 & 0.864 & -0.001 & [-0.387, 0.164] & [-0.263, 0.275] \\
1.2.1 & Precision Drop & -0.138 & 0.312 & 0.019 &  0.012 & 0.930 & -0.009 & 0.926 & -0.004 & [-0.402, 0.157] & [-0.254, 0.285] \\
1.2.1 & Recall Drop    & -0.118 & 0.387 & 0.014 & -0.055 & 0.688 & -0.043 & 0.666 & -0.004 & [-0.373, 0.111] & [-0.326, 0.232] \\
1.2.1 & F1 Drop        & -0.141 & 0.302 & 0.020 & -0.039 & 0.774 & -0.031 & 0.744 & -0.004 & [-0.406, 0.143] & [-0.330, 0.249] \\
\hline

\multicolumn{12}{|c|}{\textbf{Experiment 1.2.2 — Sex Label, Feature Removal}} \\
\hline
1.2.2 & Accuracy Drop  & -0.068 & 0.621 & 0.005 & -0.154 & 0.258 & -0.105 & 0.286 & -0.001 & [-0.344, 0.202] & [-0.429, 0.132] \\
1.2.2 & Precision Drop & -0.026 & 0.850 & 0.001 & -0.141 & 0.301 & -0.093 & 0.328 & -0.000 & [-0.207, 0.204] & [-0.397, 0.135] \\
1.2.2 & Recall Drop    & -0.053 & 0.700 & 0.003 & -0.055 & 0.690 & -0.038 & 0.702 & -0.000 & [-0.289, 0.192] & [-0.318, 0.221] \\
1.2.2 & F1 Drop        & -0.071 & 0.606 & 0.005 & -0.150 & 0.271 & -0.097 & 0.307 & -0.000 & [-0.368, 0.220] & [-0.415, 0.142] \\
\hline

\multicolumn{12}{|c|}{\textbf{Experiment 2.1 — Learning Curve Slope}} \\
\hline
2.1 & LC Slope & -0.115 & 0.099 & 0.013 & -0.185 & 0.007 & -0.129 & 0.006 & -0.042 & [-0.274, 0.049] & [-0.326, -0.031] \\
\hline

\multicolumn{12}{|c|}{\textbf{Experiment 2.2 — Train–Validation Gap Slope}} \\
\hline
2.2 & Gap Slope & 0.065 & 0.349 & 0.004 & 0.103 & 0.138 & 0.072 & 0.124 & 0.000 & [-0.097, 0.206] & [-0.049, 0.236] \\
\hline

\end{tabular}
}
\end{table*}

\begin{table*}[t]
\centering
\caption{Complete Summary of Inferential Metrics for Multivariate Effect Size Experiments — Adult Census Dataset}
\label{table:multivariate_adult_summary}
\resizebox{0.9\textwidth}{!}{
\begin{tabular}{|c|c|c|c|c|c|c|c|c|c|c|c|}
\hline
Exp. & Metric &
Pearson $r$ & $p$ & $R^2$ &
Spearman $\rho$ & $p$ &
Kendall $\tau$ & $p$ &
Slope &
Pearson CI &
Spearman CI \\
\hline

\multicolumn{12}{|c|}{\textbf{Experiment 1.1.1 — Income Label, Multivariate Effect Size}} \\
\hline
1.1.1 & Accuracy  & 0.095 & 0.123 & 0.009 & 0.116 & 0.060 & 0.083 & 0.054 & 0.056 & [-0.073, 0.250] & [-0.004, 0.244] \\
1.1.1 & Precision & 0.117 & 0.058 & 0.014 & 0.050 & 0.418 & 0.035 & 0.403 & 0.209 & [-0.175, 0.343] & [-0.083, 0.180] \\
1.1.1 & Recall    & -0.109 & 0.076 & 0.012 & 0.074 & 0.228 & 0.053 & 0.204 & -0.187 & [-0.314, 0.140] & [-0.056, 0.196] \\
1.1.1 & F1 Score  & -0.117 & 0.058 & 0.014 & 0.081 & 0.188 & 0.056 & 0.180 & -0.169 & [-0.355, 0.170] & [-0.042, 0.208] \\
\hline

\multicolumn{12}{|c|}{\textbf{Experiment 1.1.2 — Sex Label, Multivariate Effect Size}} \\
\hline
1.1.2 & Accuracy  & 0.284 & $2.79\times10^{-6}$ & 0.081 & 0.345 & $8.93\times10^{-9}$ & 0.235 & $3.73\times10^{-8}$ & 0.168 & [0.144, 0.430] & [0.237, 0.444] \\
1.1.2 & Precision & 0.209 & $6.51\times10^{-4}$ & 0.043 & 0.245 & $5.79\times10^{-5}$ & 0.167 & $6.34\times10^{-5}$ & 0.109 & [0.092, 0.329] & [0.132, 0.355] \\
1.1.2 & Recall    & 0.311 & $2.62\times10^{-7}$ & 0.096 & 0.330 & $3.95\times10^{-8}$ & 0.225 & $7.21\times10^{-8}$ & 0.197 & [0.196, 0.421] & [0.214, 0.438] \\
1.1.2 & F1 Score  & 0.339 & $1.63\times10^{-8}$ & 0.115 & 0.381 & $1.56\times10^{-10}$ & 0.255 & $9.53\times10^{-10}$ & 0.152 & [0.217, 0.458] & [0.274, 0.477] \\
\hline

\multicolumn{12}{|c|}{\textbf{Experiment 1.2.1 — Income Label, Feature Removal}} \\
\hline
1.2.1 & Accuracy Drop  & 0.081 & 0.553 & 0.007 & 0.034 & 0.804 & 0.017 & 0.864 & 0.043 & [-0.074, 0.265] & [-0.233, 0.289] \\
1.2.1 & Precision Drop & 0.109 & 0.422 & 0.012 & 0.054 & 0.692 & 0.034 & 0.717 & 0.143 & [-0.041, 0.284] & [-0.220, 0.314] \\
1.2.1 & Recall Drop    & 0.135 & 0.319 & 0.018 & 0.158 & 0.244 & 0.120 & 0.232 & 0.195 & [-0.071, 0.332] & [-0.112, 0.411] \\
1.2.1 & F1 Drop        & 0.143 & 0.293 & 0.020 & 0.134 & 0.324 & 0.100 & 0.294 & 0.178 & [-0.018, 0.316] & [-0.132, 0.389] \\
\hline

\multicolumn{12}{|c|}{\textbf{Experiment 1.2.2 — Sex Label, Feature Removal}} \\
\hline
1.2.2 & Accuracy Drop  & 0.536 & $2.0\times10^{-5}$ & 0.288 & 0.067 & 0.621 & 0.042 & 0.673 & 0.083 & [0.104, 0.768] & [-0.205, 0.361] \\
1.2.2 & Precision Drop & 0.549 & $1.2\times10^{-5}$ & 0.301 & 0.119 & 0.383 & 0.085 & 0.371 & 0.095 & [0.095, 0.798] & [-0.165, 0.386] \\
1.2.2 & Recall Drop    & -0.018 & 0.895 & 0.000 & -0.129 & 0.344 & -0.106 & 0.289 & -0.003 & [-0.547, 0.523] & [-0.409, 0.179] \\
1.2.2 & F1 Drop        & 0.457 & $3.9\times10^{-4}$ & 0.209 & 0.029 & 0.830 & 0.013 & 0.887 & 0.047 & [0.022, 0.703] & [-0.241, 0.327] \\
\hline

\end{tabular}
}
\end{table*}

\begin{table*}[t]
\centering
\caption{Complete Summary of Inferential Metrics for Multivariate Effect Size Experiments — Bank Marketing Dataset}
\label{table:multivariate_bank_summary}
\resizebox{0.9\textwidth}{!}{
\begin{tabular}{|c|c|c|c|c|c|c|c|c|c|c|c|}
\hline
Exp. & Metric &
Pearson $r$ & $p$ & $R^2$ &
Spearman $\rho$ & $p$ &
Kendall $\tau$ & $p$ &
Slope &
Pearson CI &
Spearman CI \\
\hline

\multicolumn{12}{|c|}{\textbf{Experiment 1.1.1 — Term Deposit Label}} \\
\hline
1.1.1 & Accuracy  & 0.314 & $9.46\times10^{-10}$ & 0.098 & 0.294 & $1.03\times10^{-8}$ & 0.212 & $8.55\times10^{-9}$ & 0.076 & [0.210, 0.407] & [0.196, 0.384] \\
1.1.1 & Precision & 0.197 & $1.50\times10^{-4}$ & 0.039 & 0.174 & $8.54\times10^{-4}$ & 0.121 & $7.71\times10^{-4}$ & 0.346 & [0.075, 0.312] & [0.065, 0.274] \\
1.1.1 & Recall    & 0.249 & $1.45\times10^{-6}$ & 0.062 & 0.230 & $9.40\times10^{-6}$ & 0.160 & $8.59\times10^{-6}$ & 0.285 & [0.166, 0.336] & [0.135, 0.322] \\
1.1.1 & F1 Score  & 0.307 & $2.09\times10^{-9}$ & 0.095 & 0.266 & $2.54\times10^{-7}$ & 0.183 & $2.91\times10^{-7}$ & 0.346 & [0.220, 0.395] & [0.175, 0.366] \\
\hline

\multicolumn{12}{|c|}{\textbf{Experiment 1.1.2 — Housing Label}} \\
\hline
1.1.2 & Accuracy  & 0.212 & $4.71\times10^{-5}$ & 0.045 & 0.238 & $4.36\times10^{-6}$ & 0.164 & $5.53\times10^{-6}$ & 0.166 & [0.117, 0.300] & [0.143, 0.329] \\
1.1.2 & Precision & 0.109 & 0.037 & 0.012 & 0.120 & 0.022 & 0.083 & 0.019 & 0.082 & [0.007, 0.202] & [0.021, 0.219] \\
1.1.2 & Recall    & 0.268 & $2.01\times10^{-7}$ & 0.072 & 0.276 & $8.90\times10^{-8}$ & 0.184 & $1.98\times10^{-7}$ & 0.292 & [0.176, 0.350] & [0.179, 0.363] \\
1.1.2 & F1 Score  & 0.234 & $6.15\times10^{-6}$ & 0.055 & 0.246 & $1.99\times10^{-6}$ & 0.165 & $3.20\times10^{-6}$ & 0.183 & [0.140, 0.322] & [0.147, 0.335] \\
\hline

\multicolumn{12}{|c|}{\textbf{Experiment 1.2.1 — Term Deposit Label, Feature Removal}} \\
\hline
1.2.1 & Accuracy Drop  & 0.343 & 0.005 & 0.118 & 0.189 & 0.135 & 0.148 & 0.116 & 0.054 & [-0.091, 0.692] & [-0.072, 0.424] \\
1.2.1 & Precision Drop & -0.055 & 0.665 & 0.003 & 0.199 & 0.115 & 0.146 & 0.111 & -0.091 & [-0.605, 0.373] & [-0.075, 0.431] \\
1.2.1 & Recall Drop    & 0.148 & 0.243 & 0.022 & 0.093 & 0.467 & 0.073 & 0.439 & 0.110 & [-0.027, 0.322] & [-0.159, 0.340] \\
1.2.1 & F1 Drop        & 0.184 & 0.146 & 0.034 & 0.155 & 0.220 & 0.111 & 0.228 & 0.174 & [-0.007, 0.381] & [-0.102, 0.386] \\
\hline

\end{tabular}
}
\end{table*}

\subsection{Experiment 1}


\textbf{Experiment 1.1.1:}

\textit{Average Effect Size - Adult Census Dataset:}
The Pearson r-values do not reveal any relationship and the corresponding p-values indicate that any relationship that may exist is statistically insignificant. Figure \ref{fig:exp111_income} is consistent with this result and shows no linear trend in the data. The Spearman $\rho$ value for recall suggests a very weak positive monotonic relationship. The corresponding p-value indicates that although the relationship is weak, it is unlikely to occur by chance. The Spearman for the other performance metrics does not reveal any statistically detectable relationships. The slope for recall is approximately 10\% and the Spearman confidence interval for recall excludes zero. These results align with the positive monotonic relationship observed and confirm consistency. 

\textit{Multivariate Effect Size - Adult Census Dataset:}
The metrics in this experiment suggest no statistically significant relationship between multivariate effect size and performance.

\textit{Multivariate Effect Size - Bank Dataset:}
The Pearson r-values in Table \ref{table:multivariate_bank_summary} demonstrate a moderate positive relationship and are considered significant based on the extremely small p-values. Figure \ref{fig:exp111_deposit}, the Spearman $\rho$, and the Kendall $\tau$ align, revealing a weak monotonic relationship. The Pearson and Spearman confidence intervals confirm these trends are consistent across runs. 

\textbf{Experiment 1.1.2:}

\textit{Average Effect Size - Adult Census Dataset:}
The Pearson r-values and corresponding p-values indicate that a linear negative relationship is statistically detectable, but is very weak. As the absolute average effect size increases the performance slightly decreases. In Accuracy and F1-Score this relationship is slightly stronger. The Spearman $\rho$ values for Accuracy, Precision, and F1-Score suggest a very weak negative monotonic relationship. And the corresponding p-values for these indicate that although the relationship is weak, it is likely not a random result. The Kendall $\tau$ and slope values are quite small, but they do align with the negative trend observed. In the confidence intervals, the Pearson excludes zero for all performance metrics. Further confirming that although the relationship is weak, it is consistent. Similarly with Spearman confidence intervals, zero is excluded for Accuracy and F1-Score. Figure \ref{fig:exp112_gender} visually reflects these results as there is a very slight downtrend when compared to Figure \ref{fig:exp111_income}.

\textit{Multivariate Effect size - Adult Census Dataset:}
On the contrary, the Pearson r-values in the experiment that utilized the Mahalanobis separation indicates a weak to moderate positive relationship. This relationship is considered statistically significant and is further validated by the monotonic metrics and Figure \ref{fig:exp112_sex}.

\textit{Multivariate Effect Size - Bank Dataset:}
Here, a similar, but slightly weaker relationship as the one shown by the Adult Census data.

\textbf{Experiment 1.2.1:}

\textit{Effect Size Per Feature (Adult Census Dataset):}
The Pearson r-values suggest a weak negative correlation, but corresponding p-values are very high. This indicates that statistically we cannot reject the null hypothesis. Figure \ref{fig:exp121_income} and the remaining metrics do not suggest any statistically detectable relationships between the effect size of the removed feature and the change in performance.

\textit{Multivariate Effect Size Drop (Adult Census Dataset):}
The metrics in this experiment suggest no statistically significant relationship between multivariate effect size drop and performance drop.

\textit{Multivariate Effect Size Drop (Bank Dataset):}
The only slightly significant trend is a moderate positive relationship revealed by the Pearson r-value for accuracy. Overall the results provide no evidence for a detectable relationship in this experiment.

\textbf{Experiment 1.2.2 - Effect Size Per Feature (Adult Census Dataset):}

\textit{Effect Size Per Feature (Adult Census Dataset):}
The Pearson r-values and Kendall values show no relationship. The Spearman $\rho$-values for accuracy drop, precision drop, and F1 drop signal a weak negative monotonic relationship. However, the corresponding p-values for these three inferential metrics confirm that they are statistically insignificant. Figure \ref{fig:exp122_gender} and the confidence intervals align with these results, showing no detectable consistent relationship. 

\textit{Multivariate Effect Size Drop (Adult Census Dataset):}
There is a moderate to strong statistically significant linear relationship revealed by the Pearson r-values in Table \ref{table:multivariate_adult_summary} between the drop in multivariate effect size and accuracy drop, precision drop, and F1 drop. The rest of the metrics are statistically insignificant and weak.

\textit{Multivariate Effect Size Drop (Bank Dataset):}
The results reflect a similar but weaker trend than that of the Adult Census data.

The results for these two multivariate experiments are fragile because there are only a few influential points that define the relationship. Figure \ref{fig:exp122_sex} reveals most points have little to no change in multivariate effect size compared to the baseline.


\subsection{Experiment 2}

\textbf{Experiment 2.1:}

\textit{Average Effect Size (Adult Census Dataset):}
The Pearson r-value for the experiment indicates a negative correlation, but the p-value considers this statistically insignificant. The Spearman $\rho$-value suggests a weak negative monotonic relationship between absolute average effect size and the slope of the learning curve. The corresponding p-value confirms this relationship is unlikely due to random chance. The Kendall value and its p-value support this observation and suggest a very weak relationship as well. As absolute average effect size goes up, the slope of the learning curve slightly decreases. The Spearman confidence interval confirms consistency.

\textit{Multivariate Effect Size (Adult Census Dataset):}
There is no detectable or statistically significant relationship for this experiment. The results suggest there is no correlation between the multivariate effect size and data sufficiency in regards to the training curve.

\textit{Multivariate Effect Size (Bank Dataset):}
The results are consistent with those observed in the Adult Census dataset.

\textbf{Experiment 2.2:}

\textit{Average Effect Size (Adult Census Dataset):}
This experiment indicates no relationship between effect size and slope of the magnitude of the difference in error between the training and validation sets. The data points and metrics suggest that determining a sufficient sample size to train your data is dependent on some other factor of our dataset along with the specific model utilized. Figure \ref{fig:exp21} is consistent with the results of both experiments. On the left panel, there is a slight downtrend, though it is not prominent. The right panel reveals no observable relationship.

\textit{Multivariate Effect Size (Adult Census Dataset):}
The results align with those observed from the average effect size experiment. There is no detectable or statistically significant relationship based on the results.

\textit{Multivariate Effect Size (Bank Dataset):}
The results are consistent with those observed in the Adult Census dataset. 

\section{Discussion}

 
Our findings suggest that simple, univariate summaries of class separability, even when aggregated across features via mean absolute effect size, do not reliably reflect downstream model quality or sample-size sufficiency in realistic, mixed-type tabular data. A key reason is that the “average effect size” statistic compresses a high-dimensional, heterogeneous feature set into a single scalar, discarding the structure that learning algorithms exploit. In both Adult Census and Bank, predictive signal can arise from feature interactions, nonlinearities, and multi-category variables whose class-conditional structure is not well captured by a single marginal statistic. As a result, a dataset may look weakly separable when features are examined one at a time, yet still be easy to learn because of feature interactions; alternatively, it may look strongly separable due to redundant or spurious correlations that do not generalize.

Adding a multivariate separability metric (Mahalanobis separation) provides a more interaction-aware view and, in some settings, aligns more clearly with downstream performance. In the Bank dataset row-subset experiments, multivariate separability shows a moderate positive relationship with performance and is statistically significant. However, this behavior is not universal: in feature-removal settings the evidence is weak (only slightly significant for accuracy in one case), and the multivariate trends can be fragile due to a few influential points. Moreover, multivariate separability does not meaningfully predict learning-curve convergence in either dataset, indicating that separability alone is insufficient for pre-hoc sample-size sufficiency estimation.

Additionally, the mapping from data to performance is model- and metric-dependent. Effect size is a population statistic independent of the learner, whereas generalization error depends on inductive bias, regularization, optimization, and the evaluation metric. For example, tree-based models can exploit nonlinear threshold effects and high-order feature interactions, while linear models rely more directly on marginal separability under the chosen feature representation. In this setting, a single effect-size aggregate cannot capture whether the available feature is aligned with the hypothesis class of the model family. This helps explain why correlations were small even when statistically detectable.

Taken together, our study indicates that prospective data sufficiency assessment likely requires statistics that (i) respect feature interactions and representation, and (ii) incorporate at least minimal model assumptions. Two practical directions are promising. First, multivariate measures are more promising than univariate aggregation, but our results suggest they can be dataset- and setting-dependent and should be paired with stability checks (e.g., resampling sensitivity) and complementary predictors. Second, adopt model-aware pre-hoc predictors that are still inexpensive relative to full training. These approaches preserve the goal of prospective assessment while acknowledging that “data quality” is inseparable from the learner and representation used. As extensions, future work can explore representation-based metrics (e.g., separability measured in learned feature spaces or via simple probes) which may capture interaction structure beyond second-order multivariate summaries while remaining practical.

More broadly, our results suggest that while simple descriptive statistics are useful for understanding marginal feature behavior, they are not reliable stand-alone predictors of model performance or sample-size sufficiency.

\section{Conclusion}

This paper investigated whether simple, pre-training measures of class separability can serve as practical heuristics for (i) predicting downstream classification performance and (ii) anticipating sample-size sufficiency via learning-curve convergence on mixed-type tabular data. Using two UCI datasets (Adult Census and Bank), we evaluated both a univariate statistic (mean absolute feature effect size) and a multivariate separability metric (Mahalanobis separation) across two complementary settings—varying row subsets with fixed features and varying feature sets via leave-one-feature-out—and four model families (logistic regression, decision tree, random forest, and a neural network). Overall, associations between univariate effect-size aggregation and predictive metrics were consistently weak and often inconsistent. Multivariate separability showed clearer alignment with performance in some settings, but this behavior was not universal and could be fragile under feature-removal and sensitivity to influential points. Importantly, neither univariate nor multivariate separability meaningfully predicted learning-curve convergence, indicating that separability alone is insufficient as a stand-alone predictor of dataset adequacy. Future work toward data sufficiency assessment should therefore combine interaction-aware statistics with stability checks and low-cost, model-aware predictors that improve practical decision relevance for experimental design and data acquisition.

\bibliographystyle{IEEEbib}
\bibliography{aaai25}
\end{document}